\documentclass[journal]{IEEEtran}
\ifCLASSINFOpdf
 \usepackage[pdftex]{graphicx}
\else
\usepackage[dvips]{graphicx}
\fi
\usepackage{amsmath}
\ifCLASSOPTIONcompsoc
  \usepackage[caption=false,font=normalsize,labelfont=sf,textfont=sf]{subfig}
\else
  \usepackage[caption=false,font=footnotesize]{subfig}
\fi
\usepackage{textcomp}

\usepackage{dblfloatfix}
\usepackage[normalem]{ulem}

\usepackage{float}

\usepackage{multirow}
\usepackage{comment}
\usepackage{textgreek}
\usepackage{tabstackengine}
\begin{document}
%
\title{Multi-Axis Force Sensing in Robotic Minimally Invasive Surgery With No Instrument Modification } 
%
%
%

\author{Amir Hossein~Hadi Hosseinabadi,~\IEEEmembership{Student Member,~IEEE,}
        Septimiu E.~Salcudean,~\IEEEmembership{Fellow,~IEEE}
\thanks{Amir Hossein Hadi Hosseiabadi and Septimiu Salcudean are with the Department
of Electrical and Computer Engineering, University of British Columbia,
BC, Canada  e-mail: ahhadi@ece.ubc.ca, tims@ece.ubc.ca.}
\thanks{Support for this Research was provided by NSERC, CFI and the C.A. Laszlo Chair held by Prof. Salcudean.}
\thanks{Manuscript received ???.}}

%
%

\markboth{IEEE Transactions on Robotics,~Vol.~?, No.~?, March~2021}%
{Shell \MakeLowercase{\textit{{\em et al.}}}: Bare Demo of IEEEtran.cls for IEEE Journals}
%



\maketitle

\begin{abstract}
This paper presents a novel multi-axis force-sensing approach in robotic minimally invasive surgery with no modification to the surgical instrument. Thus, it is adaptable to different surgical instruments. A novel 6-axis optical force sensor, with local signal conditioning and digital electronics,  was mounted onto the proximal shaft of a da Vinci EndoWrist instrument. A new cannula design comprising an inner tube and an outer tube was proposed. The inner tube is attached to the cannula’s interface to the robot base through a compliant leaf spring with adjustable stiffness. It allows bending of the instrument shaft due to the tip forces. The outer tube mechanically filters out the body forces from affecting the instrument's bending behavior. A mathematical model of the sensing principle was developed and used for model-based calibration. A data-driven calibration based on a shallow neural network architecture comprising a single 5-nodes hidden layer and a 5$\times$1 output layer is discussed. Extensive testing was conducted to validate that the sensor can successfully measure the lateral forces and moments and the axial torque applied to the instrument's distal end within the desired resolution, accuracy, and range requirements.
\end{abstract}

\begin{IEEEkeywords}
Force and Tactile Sensing, Surgical Robotics: Laparoscopy, Haptics and Haptic Interfaces, Telerobotics and Teleoperation, dVRK
\end{IEEEkeywords}

%
\IEEEpeerreviewmaketitle

\section{Introduction}\label{sec: introduction}
%
%
%
%
\subsection{Introduction}
\IEEEPARstart{W}{ith} the rapid developments in the fields of robotics, computer vision, and data-driven learning, Robot-Assisted Minimally Invasive Surgery (RAMIS) has been showing exponential popularity over the past decade \cite{Intuitive}. In addition to the benefits of a minimally invasive procedure such as less tissue trauma, blood loss, and faster recovery, RAMIS provides improved ergonomic factors, reducing surgeon fatigue, 3D surgical vision, automatic movement transformations, fine motions, hand tremor filtering, motion scaling, and improved instrument dexterity all of which lead to higher surgery precision \cite{Aviles2017,Bandari2020a}.

Despite the recent advancements, the clinical RAMIS systems do not provide haptic perception \cite{Abiri2017}. This deprives the surgeon of the rich information embedded in palpating the tissue and direct interaction with surgical tools. Traditionally surgeons use palpation to characterize tissue properties, detect nerves and arteries, and identify abnormalities such as lumps and tumors \cite{Hong2012}. Moreover, the surgeons rely on the sense of touch to regulate the applied forces. Excessive forces can lead to tissue trauma, internal bleeding, and broken sutures. However, insufficient forces can lead to loose knots and poor sutures. Thus, many studies are targeted towards the reconstruction and evaluation of haptic feedback \cite{Hadi2019}.

Several surveys on haptic feedback and its efficacy in teleoperated robotic surgery \cite{Amirabdollahian2018,ElRassi2020}, simulation, and training \cite{Overtoom2019,Rangarajan2020} have been published in recent years. In summary, the introduction of haptic perception is proven to decrease operation time, facilitate training, improve accuracy, and enhance patient safety in complex tasks. Additionally, force information can be used to automate robotic tasks in unstructured environments, to identify tissues in real-time, to create tissue-realistic models and simulators for training, and to perform surgical skills assessment \cite{Bandari2020a,Abdi2020}. A transparent haptic interface requires a method of force estimation or force sensing.

\subsection{Background}
The force estimation techniques use the existing sensors in the hardware (encoders, potentiometers, cameras). They categorize into model-based, data-driven, or vision-based. The model-based approaches combine analytical models \cite{Anooshahpour2014} with disturbance observers \cite{Yoon2015, Tsukamoto2014} and Kalman filters \cite{Li2016}. They have limited accuracy due to the system nonlinearities, e.g. creep, hysteresis, friction, and backlash. The data-driven approaches use supervised learning methods to train a mapping from a custom set of inputs to a set of known interaction forces \cite{Li2017-Y}. They show improved accuracy compared to the analytical methods because they do not impose assumptions on the system nonlinearities \cite{Stephens2019, Abeywardena2019}. Vision-based techniques use visual feedback (mono and stereoscopic views) as the input to the supervised-learning frameworks \cite{Aviles2017, Haouchine2018}. They are computationally expensive, provide low data rates, and can be affected by instrument occlusion, bleeding, smoke, changes in the tissue properties, lighting, and camera orientation.

Strain-gauges are compact and low-cost and provide high accuracy in Wheatstone bridge arrangements.  Strain gauges have been integrated into the instrument's base \cite{Kong2018}, proximal and distal shaft \cite{Trejos2017, Wee2016}, wrist \cite{Seibold2007}, gripper \cite{Seneci2017}, and the cannula \cite{Kim2017}. However, strain gauges require special surface preparation, adhesives, and coatings. They are fragile, sensitive to electromagnetic interference, and often require overload protection and structural modification for strain amplification.

Capacitive transducers provide high resolution in a compact form factor and have better hysteresis performance compared to the strain gauges. 
With the Capacitive to Digital Converter (CDC) chips (e.g. AD7147 from Analog devices), they have been recently integrated into the instrument's base, wrist \cite{Lee2016}, and gripper \cite{Kim2015}, with promising performances.
However, capacitive sensors have a limited range and are prone to thermal and humidity drifts. 
Furthermore, as with strain gauges, custom designs are needed for each instrument type, requiring  instrument modifications.

Optical methods use light intensity (e.g. photodiodes, phototransistors), frequency (e.g. Fiber Brag Gratings), or phase (e.g. interferometry) modulation for force measurement. The optical signal can be locally converted to electric signals, or be transferred with fibers for distal processing. Placing the electronics away from the instrument tip makes sterilization easier. The optical fibers are flexible, scalable, biocompatible, electrically passive, insensitive to electromagnetic noise, and thus MRI compatible, durable against high radiation, immune to water, corrosion-resistant, and low-cost \cite{Shahzada2016}. However, optical fibers cannot be routed into small bending radii. Additionally, The presence of small and intricate parts can make the fabrication and assembly of fiber-based sensors costly.

The Light Intensity Modulation (LIM) based sensors are vulnerable to light intensity variations due to the temperature or fiber bending. Robustness can be improved by normalizing the optical signal against the emitted power. Alternatively, a redundant strain-free fiber can be used to compensate for the effect of temperature or other sources of uncertainty. Puangmali {\em et al.} \cite{Puangmali2012} presented a 3-axis force sensor with a flexible tripod structure, a stationary reflecting surface, and a pair of transmitting and receiving fibers per axis. The sensor was integrated into the shaft of a palpation instrument. Fontanelli {\em et al.} \cite{Fontanelli2017} used four optical proximity sensors in an instrumented cannula to measure the deflection of the instrument shaft. 
The proposed concept was able to measure the lateral forces with no instrument modification. However, the sensor needs to be in the abdominal cavity, and the size of the access hole has to be larger than the cannula.

The FBG sensors are wave-length coded and insensitive to the changes in the light intensity. FBGs are very sensitive, have calibration consistency, and exhibit high SNR that provide repeatable and high-resolution strain measurements. Multiple gratings can be accommodated into one fiber simplifying the design and signal routing. Thus, they are also used in shape sensing. 
FBGs have been integrated into the instrument's base \cite{Xue2018}, distal shaft \cite{Shahzada2016}, wrist \cite{Haslinger2013}, and gripper \cite{Soltani-Zarrin2018}.
Nonetheless, FBGs require interrogators for signal processing which the commercial systems cost between \$10k to \$100k. 

Micro-Electro-Mechanical (MEM) fabrication techniques are used in the development of cost-effective, compact, fully-integrated, and monolithic sensors \cite{Gafford2014, Dai2017}. Researchers have proposed the use of piezoelectric and magnetic transducers as they do not require a power supply. However, they are suitable for dynamic load measurements. Kuang {\em et al.} \cite{Kuang2020} correlated the changes in the tri-axial acceleration ellipse of a vibrating instrument shaft to the forces applied to its tip.

\subsection{Motivation}
The focus of our research was developing a multi-axis force sensor that does not require modifications to the surgical instrument and is adaptable to different surgical tools. The literature suggests that the lateral forces should be the primary focus for an effective haptic experience, and the axial force and torsion are secondary \cite{Spiers2015}. The human just-noticeable difference  (JND)  is  10\% in the range of 0.5 to 200 N increasing to 15-27\% below 0.5 N, which can be considered as a requirement on the sensor accuracy \cite{Kim2017}. A resolution of 0.2 N over a range of $\pm$10 N was assumed for the lateral forces \cite{Trejos2017}. To the best of our knowledge, no requirements for the other DoFs have been specified. The closer the sensor is to the instrument tip, the more accurate the force sensing will be; however, the requirements on sterilizability and biocompatibility are more stringent. 
We were inspired by the \AA ngstr\"om level resolution obtained by using an optical transduction principle in bending measurement of the piezo tube in atomic force microscopy \cite{Barret1991}. A conceptual evaluation of the sensing approach was presented in \cite{Hadi2019}. This article extends the proposed concept to 5-axis force sensing (lateral forces and moments and the axial torsion) and discusses a novel cannula design for mechanically filtering the body forces.
\section{Sensing Approach}\label{sec: sensor}
A thorough explanation of the optical sensing principle was presented in \cite{Hadi2019} and \cite{Hadi2020-TRM}. 
For completeness, this section presents a concise overview.
\begin{figure}[h!]
\centering
\includegraphics[width=\linewidth]{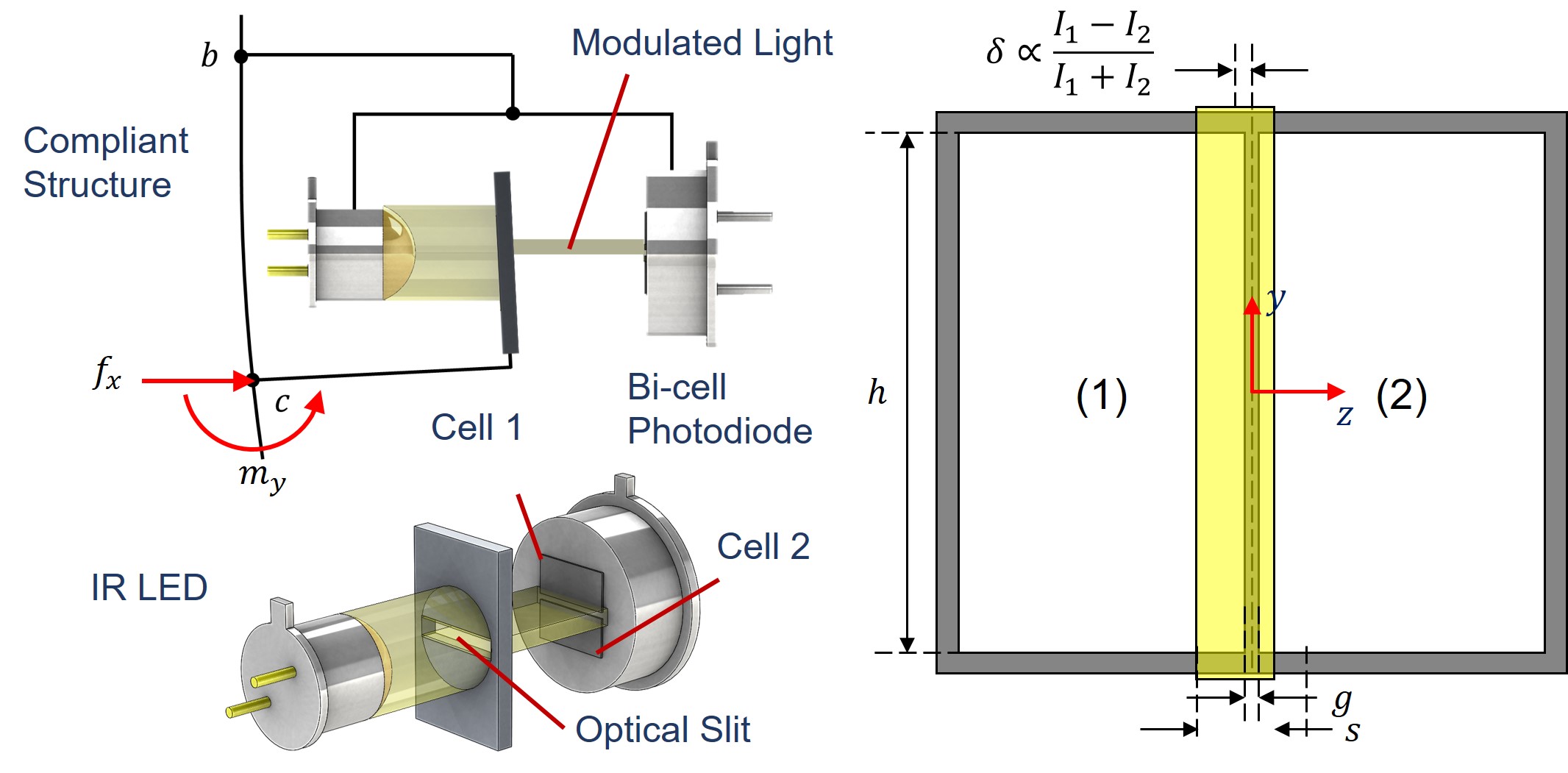}
\caption{Transduction concept. The motion of the slit w.r.t the LED-bicell pair modulates the light on the bicell's active areas.}
\label{fig: sensingConcept}
\end{figure}

The 6-axis force sensor has six sensing units in a hexagonal configuration. Every unit comprises an infrared LED normal to the cells of a bicell photodiode. An optical slit is placed in the light path (see Fig. \ref{fig: sensingConcept}) and is aligned with the gap that separates the two active surfaces of the bicell. The IR LED and the bicell are rigidly coupled and are parts of one assembly that is mounted onto a compliant structure, axially apart from the slit. When wrenches are applied, the support structure deflects and moves the slit relative to the LED-bicell pair, which modulates the light incident on each cell. It was shown that the slit displacement ($\delta$) is proportional to the normalized differential photocurrents ($n$) (\ref{eq: photocurrent}) \cite{Hadi2019}. $\kappa$ is a constant that is a function of the LED and bicell characteristics and the LED driving current, and  $c = \frac{1}{2}({s-g})$ where $s$ is the slit width and $g$ is the bicell gap width. Each module is most sensitive to displacements in the direction that is normal to the slit.
\begin{equation}
    \delta = c~n,~n = \frac{I_1-I_2}{I_1+I_2}, \,\,\,
    \textrm{  where  } \,\,
    \begin{cases}
    I_{1} = \kappa \left(1+\frac{\delta}{c} \right)\\
    I_{2} = \kappa \left(1-\frac{\delta}{c} \right)\\
    \end{cases}
    \label{eq: photocurrent}
\end{equation}

\begin{figure}[h!]
\centering
\subfloat[Exploded view]{\includegraphics[height=5cm]{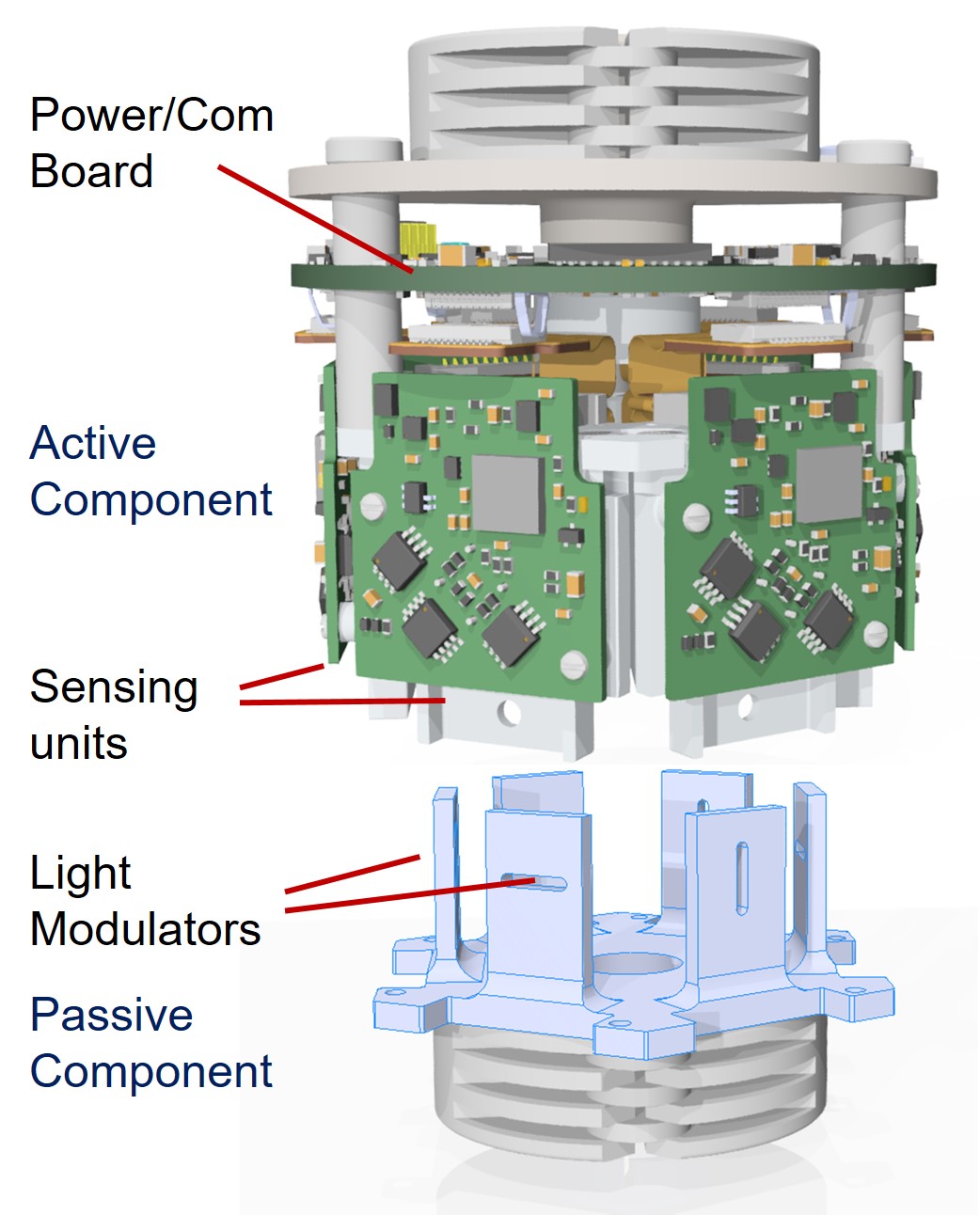}}
\hspace{0.2cm}
\subfloat[Assembled view]{\includegraphics[height=5cm]{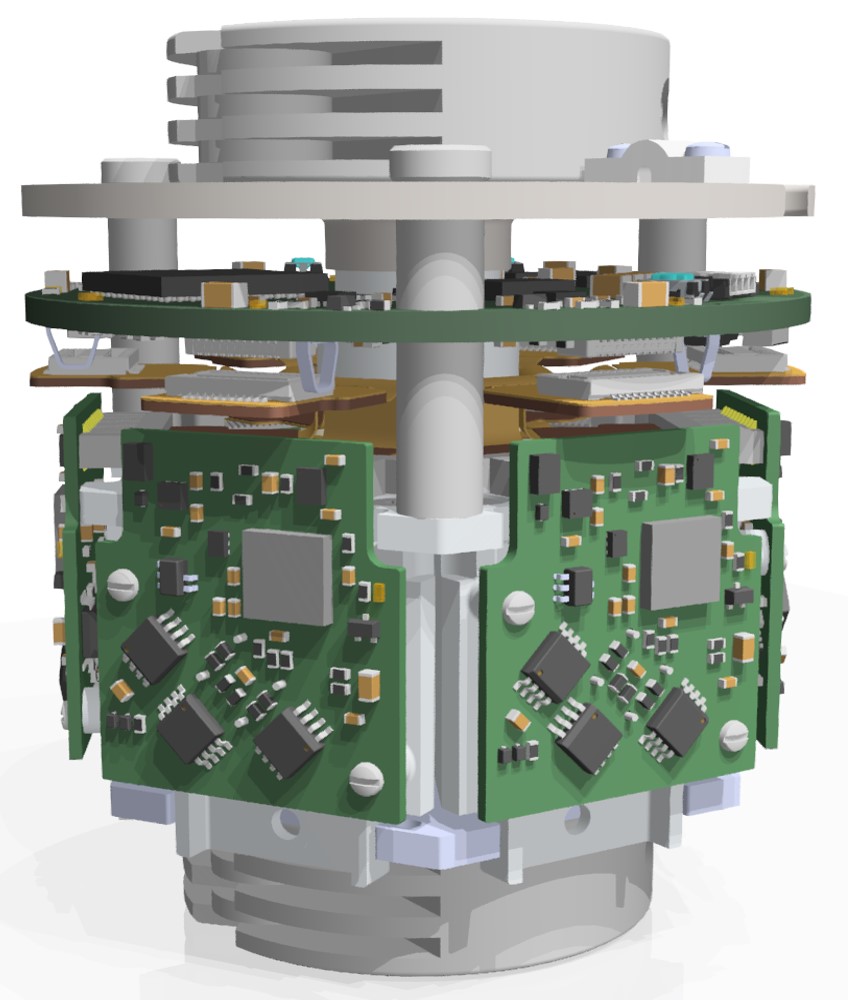}}
\caption{The exploded (a) and assembled (b) views of the 6-axis F/T sensor}
\label{fig: sensorCAD}
\end{figure}

In the sensor assembly (see Fig. \ref{fig: sensorCAD}), three units with horizontal slits are interleaved with three units with vertical slits. The horizontal slits are most sensitive to the axial force and lateral moments. However, the vertical slits are most sensitive to the lateral forces and axial torsion. The bicells have alternating orientations in accordance with the slits. The six units and all the analog and digital electronics form an active assembly. The six slits form a passive component.

A detailed explanation of the sensor's hardware and software was presented in \cite{Hadi2020-TIE}. In summary, each sensing unit has a bicell board that hosts the signal conditioning circuit with a 24-bit Analog-to-Digital Converter (ADC) for collocated sampling. The sensor has a generic Power/Com board with an onboard FPGA processor. Flexible PCBs interface the bicell boards to the Power/Com board. The sensor electronics and firmware provide an ultra-low noise rms of 2.8 \textmu V over a range of $\pm$5 V and a signal bandwidth of 500 Hz that translate into a resolution to the full-scale ratio of less than 0.0001\% and a noise PSD of 15 nV/$\sqrt{\text{Hz}}$. The sensor provides a low latency of 86 \textmu s and can achieve data rates up to 11.5 kHz making it suitable for control applications.

\begin{figure}[h!]
\centering
\includegraphics[width=0.85\linewidth]{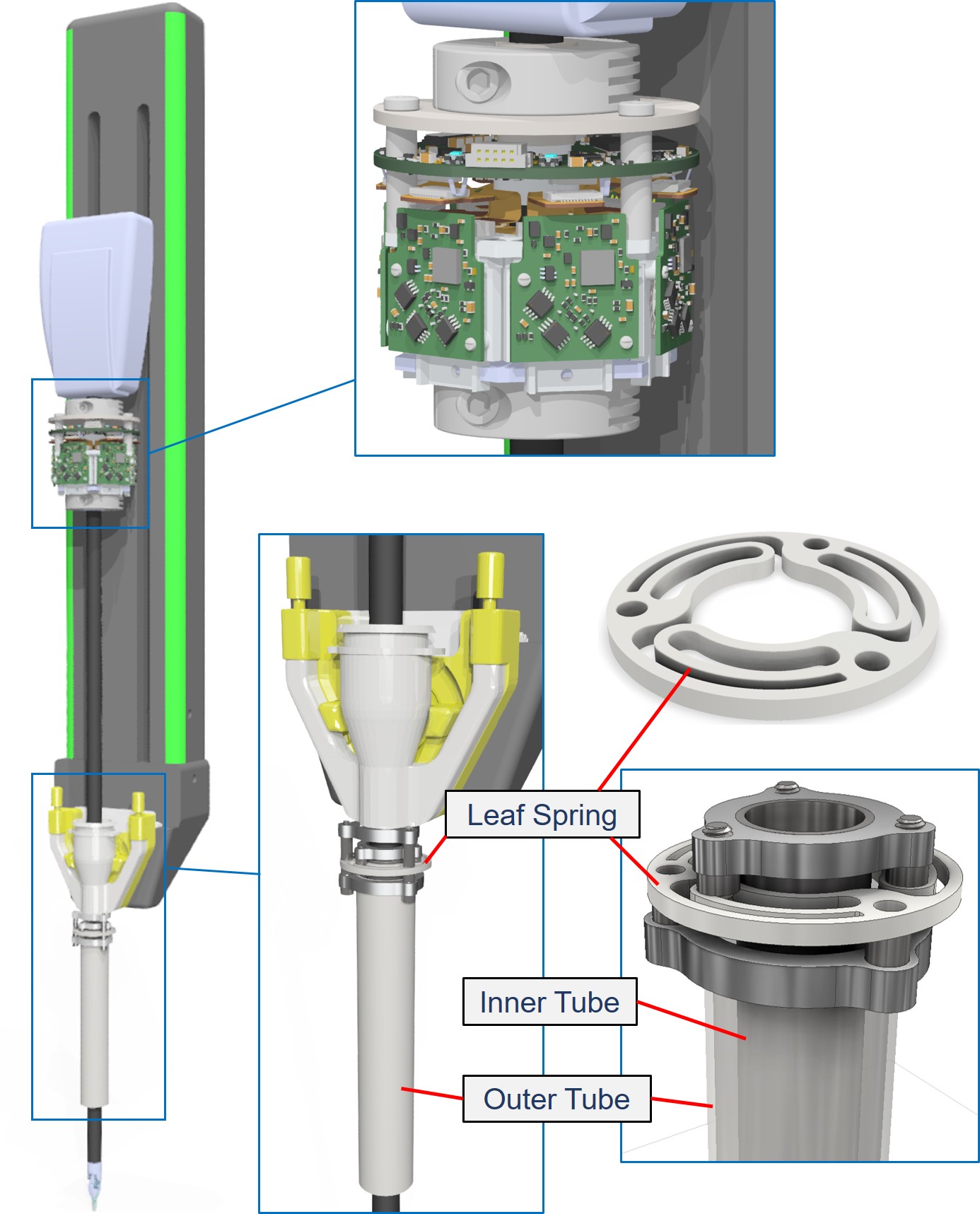}
\caption{Schematics of the proposed force sensing approach. The 6-axis optical force sensor is mounted onto the proximal shaft. The cannula is modified to have an outer tube as an overcoat.}
\label{fig: fullCAD}
\end{figure}

The optical sensor was mounted onto the proximal shaft of the instrument via screw-type mechanical clamps as shown in Fig. \ref{fig: fullCAD}. The mounting interface is not critical because it transfers no load to any of the components; therefore, it can be a set-screw connection, a mechanical, or a magnetic clamp. As explained in section \ref{sec: introduction}, force sensing at the proximal shaft can be affected by the forces between the cannula and the patient's body. To mitigate this, we modified the design of the cannula in the da Vinci classic system such that an outer tube with an outer diameter of 14.5 mm and a wall thickness of 0.5 mm covers an inner tube through which the instrument passes. The outer tube is in contact with the patient's body; thus, the body wall forces are transferred through a bolted connection to the top part of the cannula and subsequently to the robot frame. Therefore, the cannula-body forces are no longer applied to the inner tube and will have a reduced effect on the sensor measurements. This concept is similar to the overcoat method \cite{Shimachi2008}, in a more compact design.

The inner tube attaches to the top part of the cannula via a leaf spring. The leaf spring has three axis-symmetric arms with arc-shaped slots and is fabricated via water-jet out of a 1.5 mm thick spring-steel sheet. The flexible connection makes the inner tube compliant; when lateral forces or moments are applied to the instrument, it bends by pushing against the inner tube. The equivalent stiffness of the inner tube at its distal end must be high enough to prevent closing the gap (1.5 mm) between the inner tube and the outer tube at the maximum lateral forces and throughout the stroke of the instrument. The stiffness of the inner tube can be adjusted by rotating the inner tube that changes the length of the flexible arms. The proposed spring design provides uniform stiffness in all radial directions regardless of orientation about the z-axis (see Fig. \ref{fig: innerTubeComp}).

\section{Modeling}\label{sec: modeling}
\begin{figure}[h!]
\centering
\includegraphics[width=0.95\linewidth]{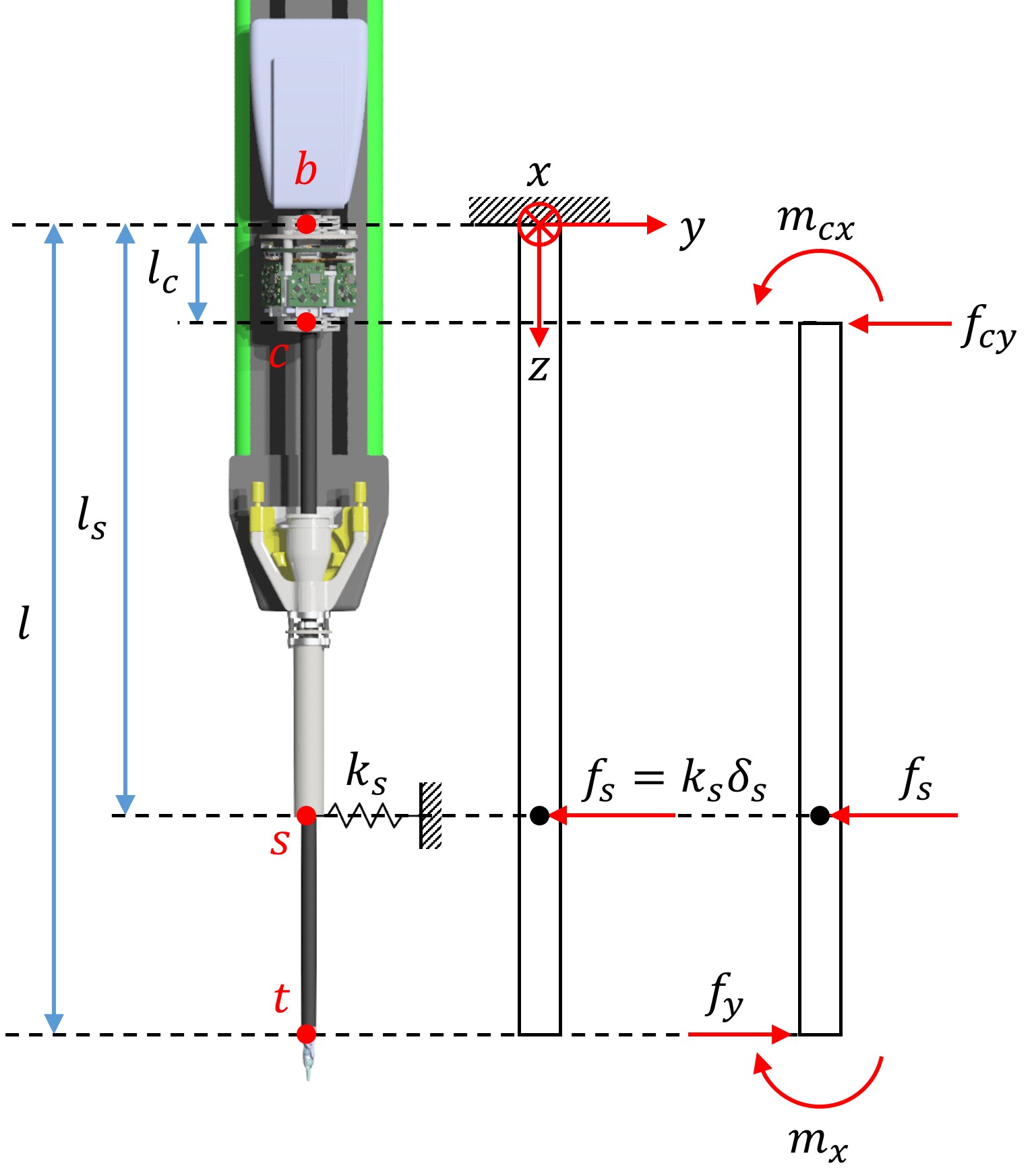}
\caption{The schematic for development of the instrument's bending model.}
\label{fig: bendModelCad}
\end{figure}

A mathematical model from the forces and moments at the distal end of the instrument shaft (point $t$ in Fig. \ref{fig: bendModelCad}) to the normalized transducers signals of the optical force sensor was developed. The mechanical constraints of the instrument shaft, which affect its bending behavior, vary as a function of the instrument's insertion into the cannula. The model was developed in two steps to decouple the sensor behavior from the changes in the boundary condition. The first part (see (\ref{eq: bendModel})) explains the sensor response to the forces applied at the clamping point of the passive component (point $c$). The second part (see (\ref{eq: varyBoundCond}) to (\ref{eq: armStiffness})) focuses on the reflected wrench vector at the shaft's cross-section at point $c$ as a function of the wrenches applied at the point $t$ considering the change in the instrument;s insertion into the cannula ($l_s$).

From geometric algebra, the principles of continuum mechanics, and the electro-optical conversion in (\ref{eq: photocurrent}) the transformation from the wrench vector at point $c$ ($\vec{w_c} = \begin{bmatrix} f_{cx},f_{cy},f_{cz},m_{cx},m_{cy},m_{cz} \end{bmatrix}^{T}$) to the vector of normalized transducers signals ($\vec{n} = \begin{bmatrix} n_1,\hdots, n_6\end{bmatrix}^{T}$) is \cite{Hadi2020-TRM}:
\begin{equation}
    \label{eq: bendModel}
    \vec{n} = C_m \vec{w_c},~~C_m = \frac{2}{c}H_G H_w,
\end{equation}
where $H_G$ is a geometric transformation matrix from the tri-axial displacement and rotation vector at point $c$ to the normal and in-plane slits displacements. $H_G$ relates to the hexagonal sensor configuration. $H_w$ is a transformation matrix from the wrenches applied at point $c$ to its tri-axial displacements and rotations. It is a function of the shaft's continuum properties.

The wrenches at point $t$ ($\vec{w_t}=\begin{bmatrix}f_{x},f_{y},f_{z},m_{x},m_{y},m_{z}\end{bmatrix}^{T}$) generate reaction forces and moments in the cross-section of the shaft at point $c$ as given in (\ref{eq: varyBoundCond}). $l$, $l_s$, and $l_c$ are the distances from the shaft's distal end ($t$), the cannula's distal end ($s$), and the clamping point of the passive component ($c$), to the clamping point of the active component ($b$), respectively (see Fig. \ref{fig: bendModelCad}).
\begin{eqnarray}
\label{eq: varyBoundCond}
    \vec{w_c} = H_c \vec{w_t} \hspace{2cm} \nonumber \\
    H_c = \begin{bmatrix}
    H_{11} & 0 & 0 & 0 & -3g(c_y) & 0 \\
    0 & H_{22} & 0 & 3g(c_x) & 0 & 0 \\
    0 & 0 & 1 & 0 & 0 & 0 \\
    0 & H_{42} & 0 & H_{44} & 0 & 0 \\
    H_{51} & 0 & 0 & 0 & H_{55} & 0 \\
    0 & 0 & 0 & 0 & 0 & 1 
    \end{bmatrix} \nonumber \\
    \left\{
    \begin{array}{l}
         H_{11} = 1 - (3l-l_s)g(c_y) \\
         H_{22} = 1 - (3l-l_s)g(c_x) \\
         H_{42} = (3l-l_s)(l_s-l_c)g(c_x)-(l-l_c) \\
         H_{44} = 1-3(l_s-l_c)g(c_x) \\
         H_{51} = (l-l_c) - (3l-l_s)(l_s-l_c)g(c_y) \\
         H_{55} = 1-3(l_s-l_c)g(c_y)
    \end{array}
    \right.\nonumber \\
    g(c) = \frac{l_s^2}{c+2l_s^3},~c_x = \frac{6EI_{xx}}{k_s},~c_y = \frac{6EI_{yy}}{k_s}.
\end{eqnarray}

\begin{figure}[h!]
\centering
\includegraphics[width=0.95\linewidth]{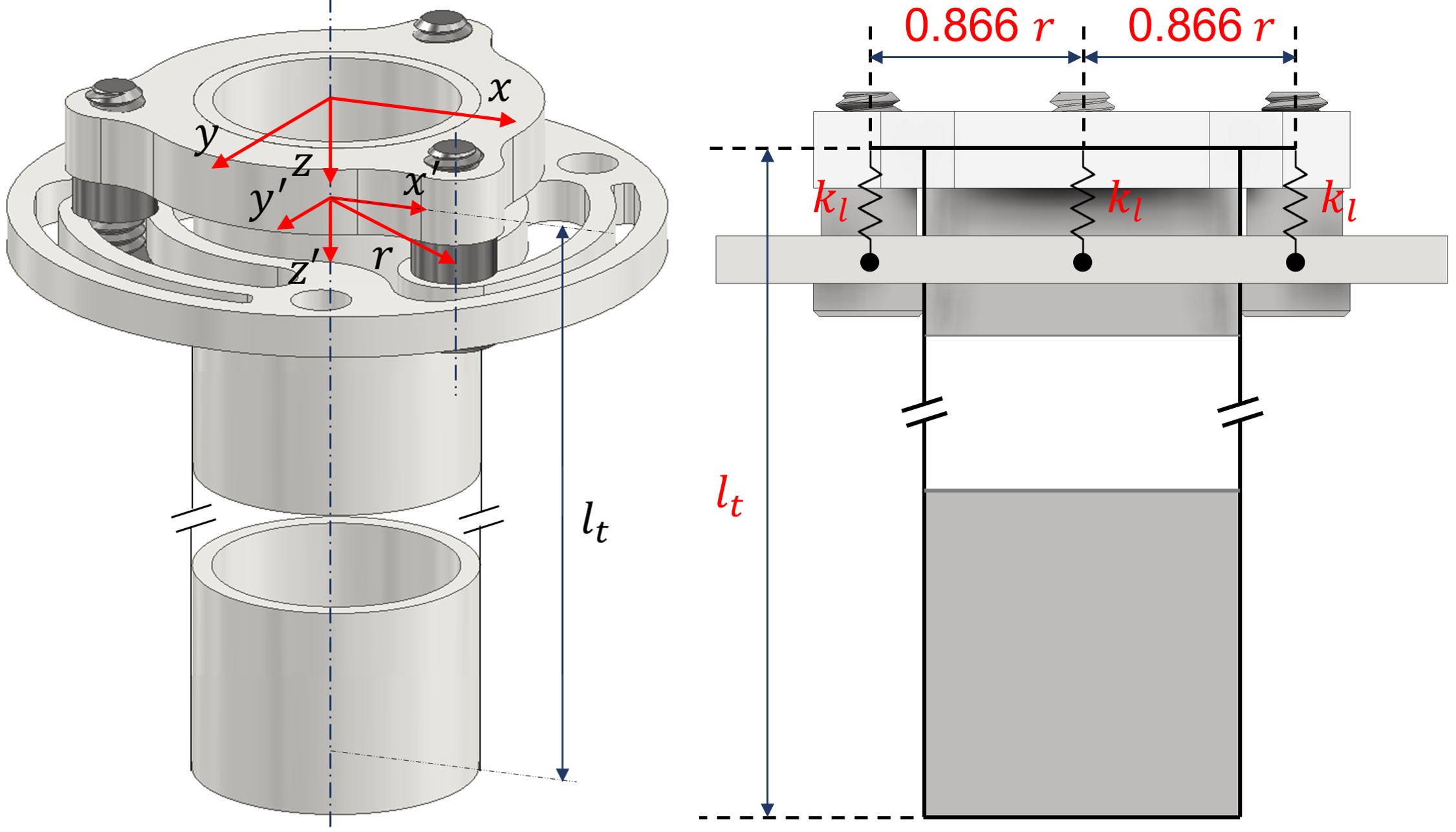}
\caption{The schematic for calculating the equivalent stiffness at the distal end of the inner tube as a function of the leaf-spring's parameters.}
\label{fig: innerTubeComp}
\end{figure}

In (\ref{eq: varyBoundCond}), $k_s$ is the cannula's equivalent stiffness at its distal end, which is a function of the stiffness of the leaf bearing's arms ($k_l$) as given in (\ref{eq: eqStif}). $r$ is the radius of the circle passing through the centerlines of the flexible arms and $l_t$ is the length of the cannula's inner tube as shown in Fig. \ref{fig: innerTubeComp}. 
\begin{equation}
    \label{eq: eqStif}
    k_s = \frac{2r}{l_t}k_l
\end{equation}
 
 $k_l$ is a function of the thickness of the spring-steel sheet, the effective length of the flexible arms, and the width of the pocketed slot. With a simplified beam bending model, $k_l$ can be approximated as:
 \begin{equation}
    \label{eq: armStiffness}
     k_l = \frac{3E_sI_s}{l_e^3},
 \end{equation}
 where $I_s = \frac{1}{12}(d_o-d_i)t^3$ with $d_o$ as the width of the arm and $d_i$ as the width of the center slot, $t$ and $E_s$ are the thickness and the modulus of elasticity of the spring-steel sheet, respectively, and  $l_e$ is the effective length of the arms. The stiffness of the flexible arms can be uniformly adjusted by rotating the inner tube and consequently changing $l_e$. Combining (\ref{eq: bendModel}) and (\ref{eq: varyBoundCond}), the normalized transducers' signals due to the applied wrench at the distal end of the instrument are:
\begin{equation}
    \vec{n} = C \vec{w_t},~~C = \frac{2}{c}H_GH_wH_c.
    \label{eq: tipToN}
\end{equation}

\section{Calibration}\label{sec: calibration}
\begin{figure}[h]
\centering
\begin{minipage}{0.345\linewidth}
\centering
\subfloat[Insrumented PSM]{\includegraphics[width=\linewidth]{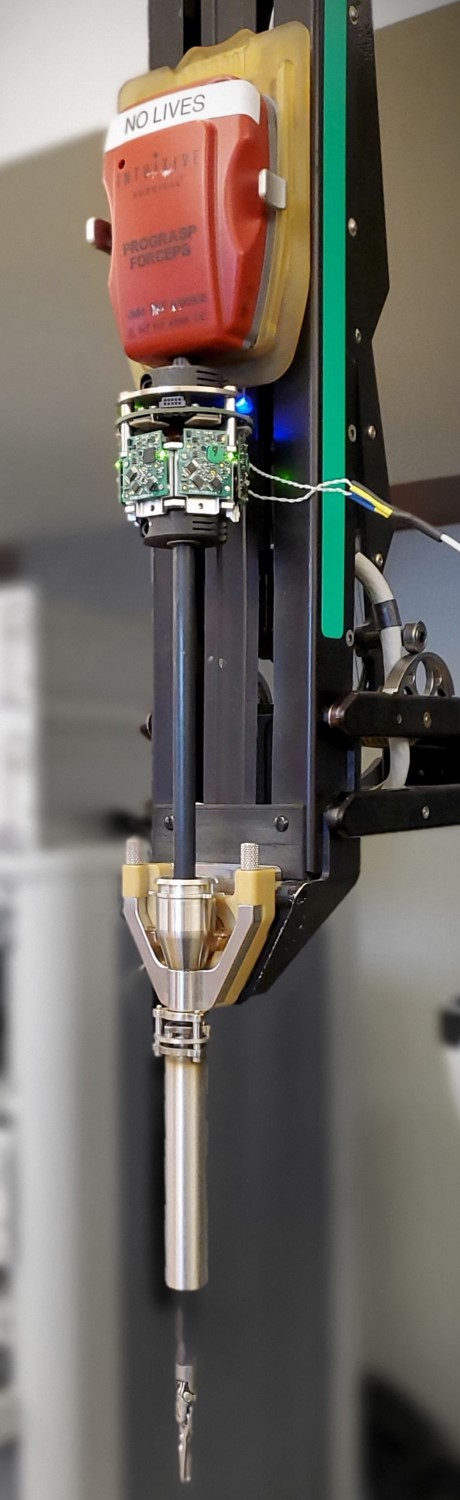}}
\end{minipage}%
\begin{minipage}{0.02\linewidth}
\hfill
\end{minipage}
\begin{minipage}{0.4\linewidth}
\centering
\subfloat[6-Axis Optical force sensor]{\includegraphics[width=\linewidth]{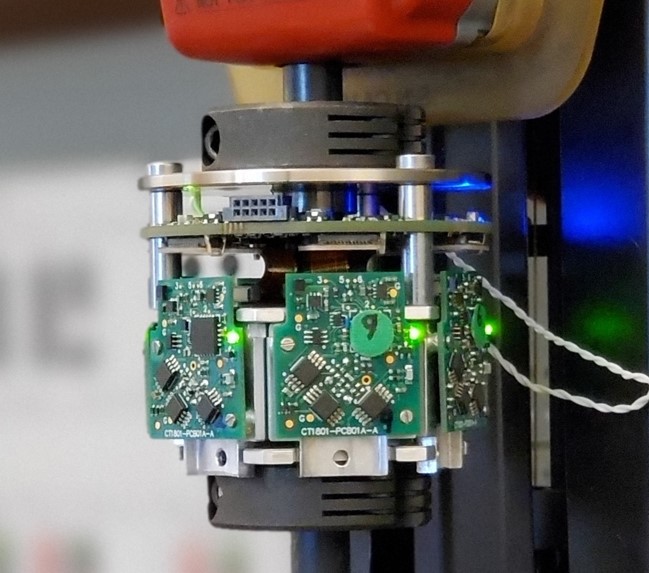}}\newline
\subfloat[Modified cannula]{\includegraphics[width=\linewidth]{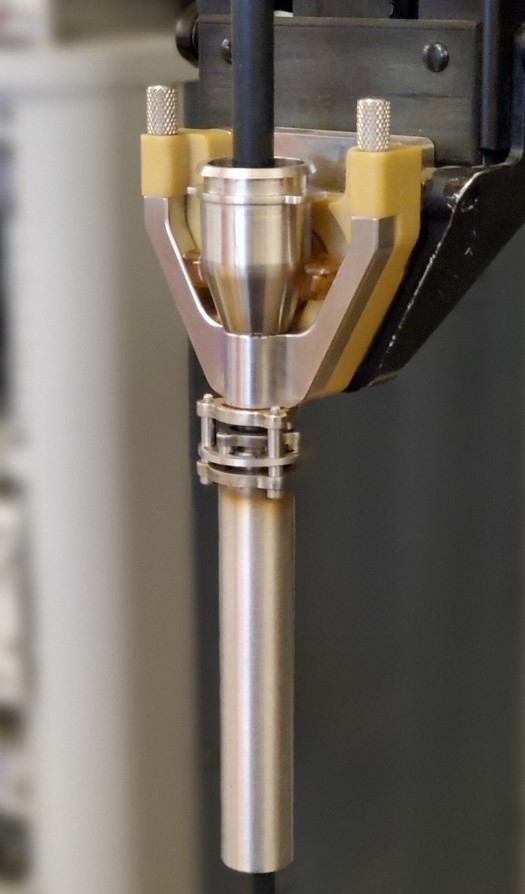}}
\end{minipage}%
\caption{The 6-axis optical force sensor with six sensing units (b) mounted onto the instrument (a). The original cannula of the standard daVinci system is replaced by the modified cannula (c).}
\label{fig: fabricatedSensor}
\end{figure}

Figure \ref{fig: fabricatedSensor} shows the optical force sensor mounted onto the proximal shaft of the ProGrasp instrument, and the modified cannula. A calibration setup (see Fig. \ref{fig: calibSetup}) were designed to measure the sensor signals throughout its insertion stroke while 6-axis forces and moments are applied to the distal end of the instrument shaft. It has a linear stage that moves synchronously with the insertion axis of the Patient Side Manipulator (PSM), and an ATI Nano43 F/T sensor as the reference. The ATI sensor is clamped to the instrument's distal shaft and attached to 3 equally-spaced radial elastic bands. The relative motion of the instrument with respect to the linear stage stretches at least one of the bands; their combination applies forces and moments to the instrument's shaft.

\subsection{Model-based}
In model-based calibration, a priori knowledge of the sensing principle in the form of an analytical model is used to map the sensor signals to a set of reference measurements by using identification methods. A validated model can be used to optimize the sensor performance and evaluate the design trade-offs; however, model-based calibration has limited accuracy because of the simplifying assumptions in model development. For example, the model explained in section \ref{sec: modeling} does not consider the friction between the instrument and the cannula and the seal, hysteresis in the shaft bending, cables creep, structural induced forces due to the wrist actuation, etc. 
\begin{figure}[h]
    \centering
    \includegraphics[width=0.9\linewidth]{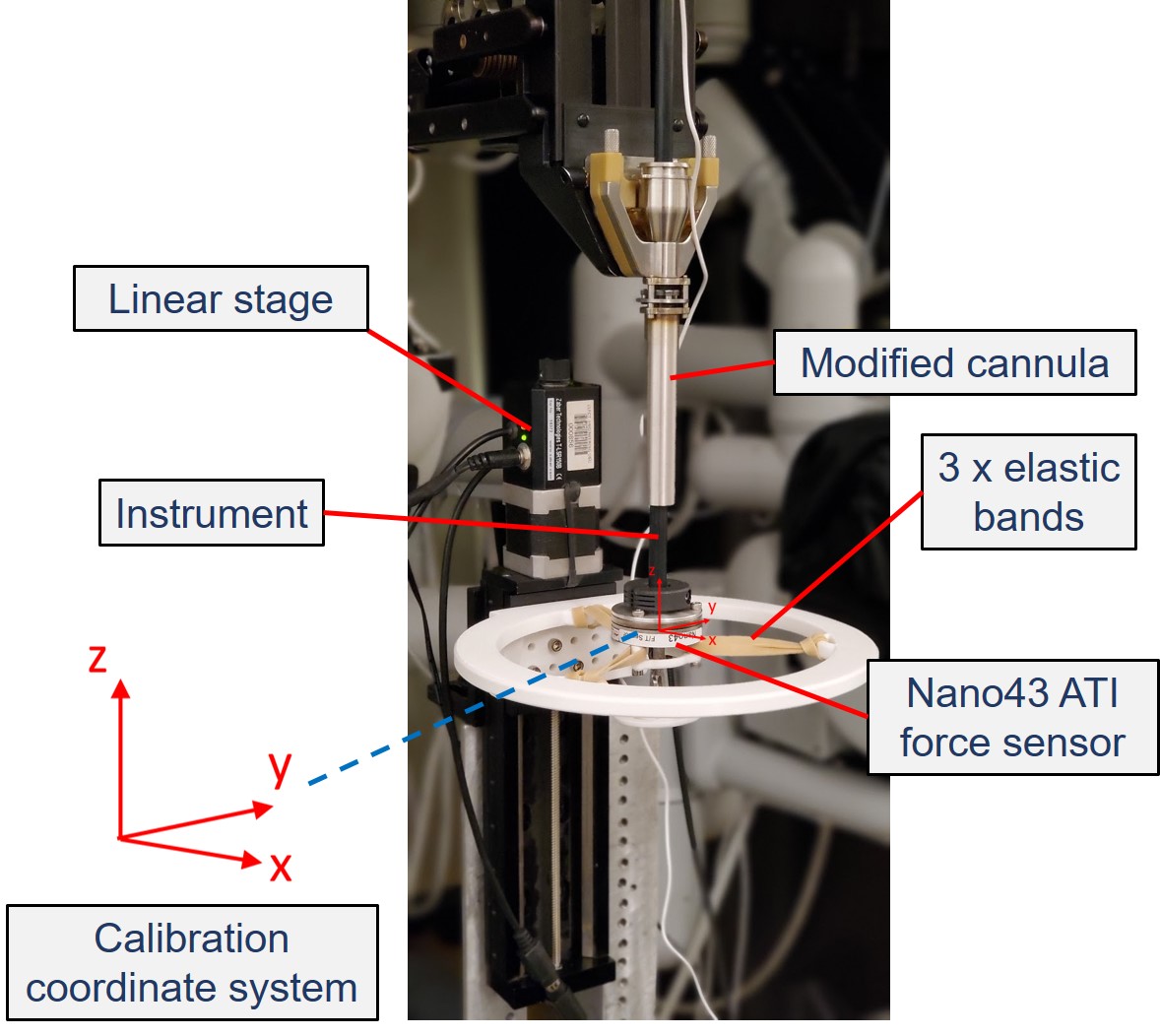}
    \caption{The calibration setup has a linear stage that moves in sync with the insertion axis of the PSM.}
    \label{fig: calibSetup}
\end{figure}
\begin{figure}[h]
    \centering
    \includegraphics[width=0.9\linewidth]{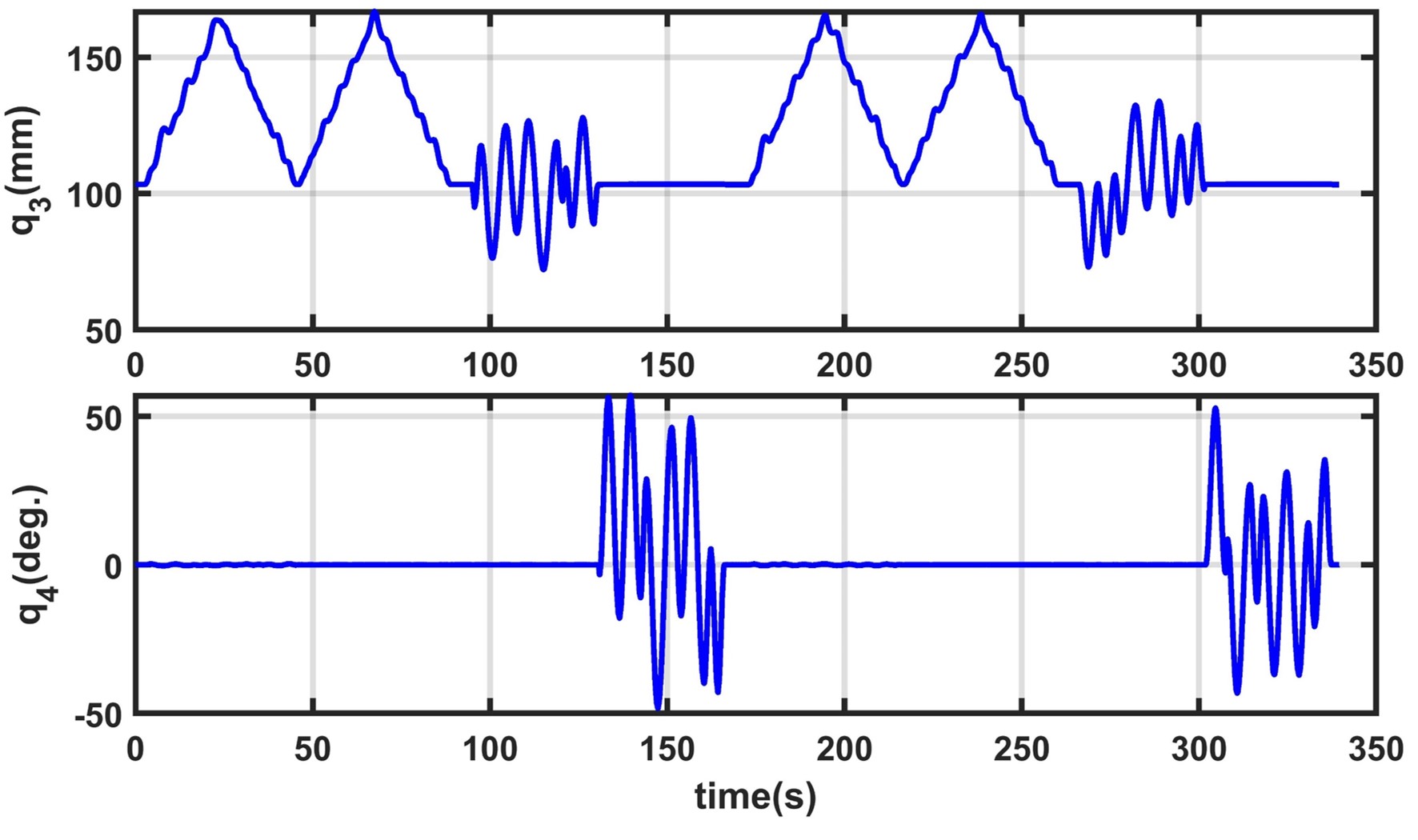}
    \caption{Instrument's displacement profile along the insertion axis ($q_3$) and axial torsion ($q_4$) in model-based calibration.}
    \label{fig: model-based motion profile}
\end{figure}
\begin{figure}[h]
    \centering
    \includegraphics[width=0.9\linewidth]{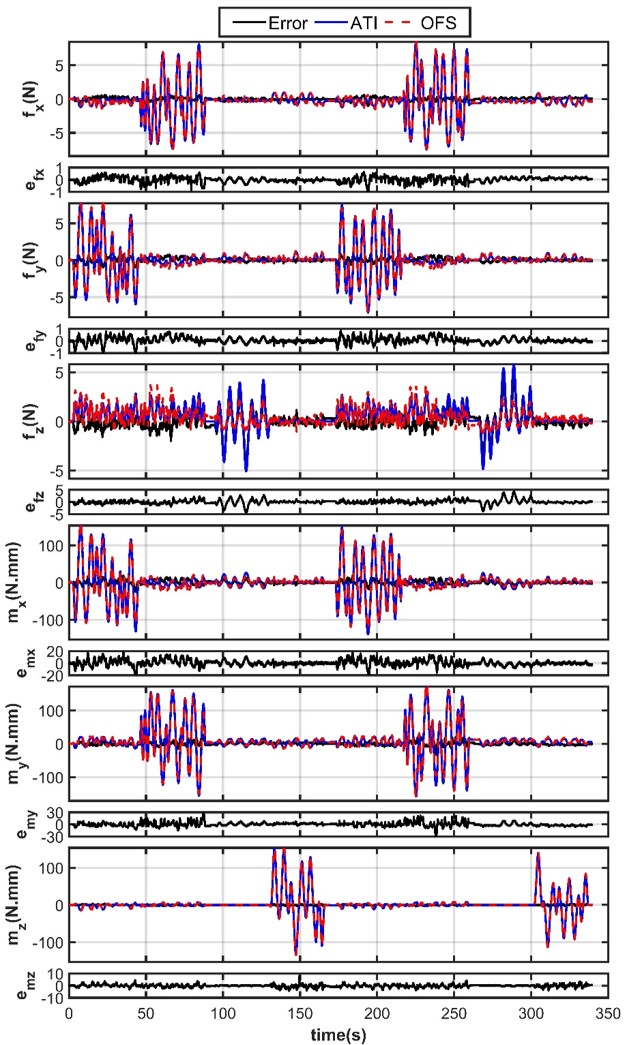}
    \caption{Force applied at the distal end of the instrument's shaft vs the calibrated optical force sensor mounted onto its proximal shaft - model-based.}
    \label{fig: model-based calibration}
\end{figure}

The setup shown in Fig. \ref{fig: calibSetup} was used for calibration. The instrument was moved axially throughout its stroke with sequentially random axial and lateral motions and torsion about its shaft.  Figure \ref{fig: model-based motion profile} shows the motion profiles of the insertion axis ($q_3$) and the axial torsion ($q_4$). The force and moments applied at the distal shaft of the instrument and measured by the ATI sensor, and the differential and common-mode signals of the optical force sensor were recorded. Two data-collection cycles were executed. The first cycle was used for calibration, and the second cycle was used for testing. The MATLAB's constrained optimization toolbox (Lsqnonlin) was used to fit a model described by (\ref{eq: tipToN}) to the measurements. From (\ref{eq: bendModel}), the term $\frac{2}{c}H_GH_w$ is a transformation from the wrench vector at point $c$ to the normalized signals of the sensing modules ($n_i$) which can be lumped into a $6\times6$ mapping matrix of $C_m$ for identification. The following constraints were defined:
\begin{equation}
    \begin{cases}
      0 < c_x < 2c_{xn} \\
      0 < c_y < 2c_{yn}\\
      l_c < l_s < l \\
      0 < l < 0.50~m
    \end{cases},
\end{equation}
where $c_{xn}$ and $c_{yn}$ are the nominal estimations by using (\ref{eq: varyBoundCond}). Given that $l_s$ decreases as the instrument penetrates into the cannula, it was reformulated as $l_s = l_{os}-q_3$ where $l_{os}$ is an offset value. $l_c$ was set to 0.035 m that is approximately the distance between points b and c (Fig.~\ref{fig: bendModelCad}). A mapping matrix ($C_m$) can be obtained for any $l_c$; therefore, if $l_c$ is not fixed, the identification does not converge. Considering the nonlinear model, the solver was executed with 1000 random initialization points within the defined constraints to ensure finding the global minimizer.

Figure \ref{fig: model-based calibration} shows the calibration results. The calibrated optical force sensor can closely reconstruct the wrench vector in all the DoFs, except the axial force. One speculation is that the axial force is affected by the friction between the instrument and the cannula when lateral forces and moments exist. The friction is not included in the model; thus, large friction forces can lead to large calibration errors in the axial direction. 

\begin{figure}[h]
    \centering
    \includegraphics[width=\linewidth]{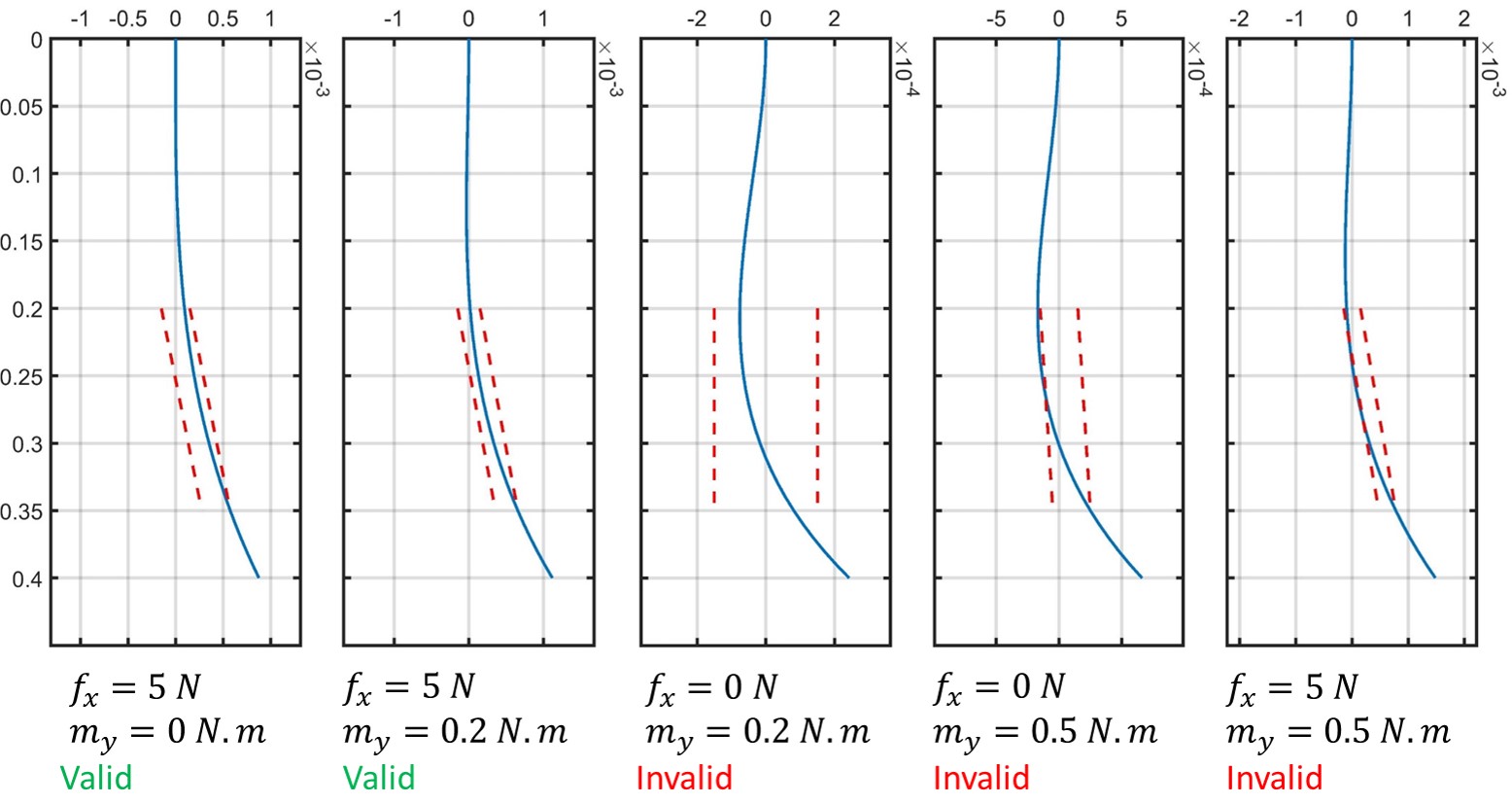}
    \caption{The dashed red line is the inner tube of the cannula and the blue line is the bending profile of the instrument shaft. This figure shows the bending scenarios where the model in (\ref{eq: tipToN}) is valid.}
    \label{fig: model-based valid}
\end{figure}

The model-based calibration is valid only for the scenarios where the modeling assumptions are valid. Figure \ref{fig: model-based valid} further elaborates on this; 1) when the moment is small, and no lateral force is applied, the deformed shape of the instrument makes no contact with the inner tube; therefore, the cannula stiffness is zero, 2) when the moment is large, the instrument hits the walls of the inner tube in addition to its tip; thus making two contact points with the cannula and the model invalid. The motion profiles in Fig. \ref{fig: model-based motion profile} was designed such that the model is valid for the range and combination of the generated wrench vector at the instrument's distal end.

\subsection{Data-driven}\label{sec: data-driven calibration}
In data-driven calibration, the sensor is considered as a black-box, and supervised learning techniques, e.g. neural networks, are adopted to identify the mapping between the sensor signals and the reference measurements. Compared to a model-based calibration, a data-driven approach is more powerful in compensating for the unmodeled nonlinearities, e.g. friction, backlash, hysteresis, changing dynamics, etc. However, it is only valid for the input measurement range and cannot be used for design optimizations.

\begin{figure}[h]
    \centering
    \includegraphics[width=0.9\linewidth]{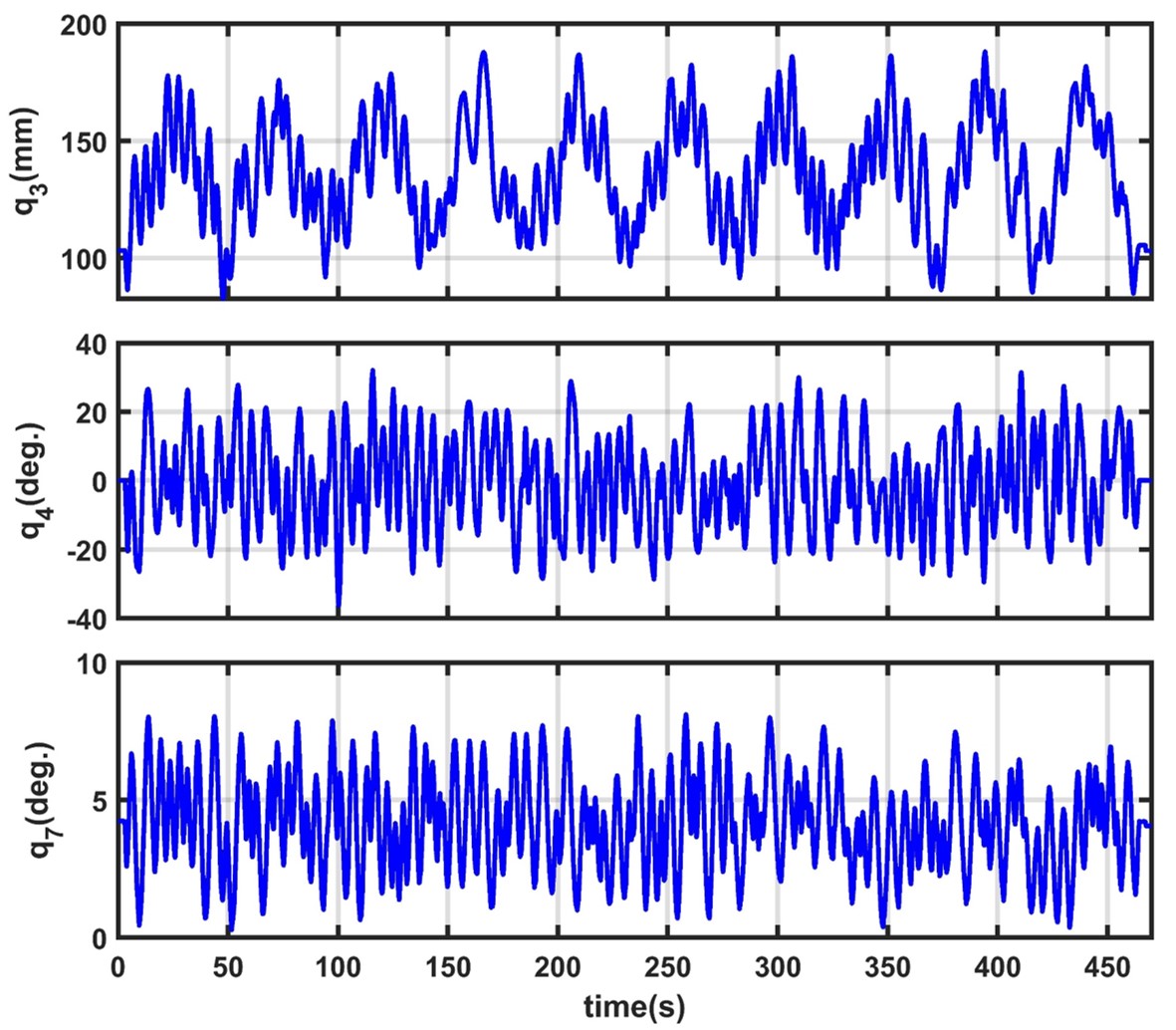}
    \caption{Instrument's displacement profile along the insertion axis ($q_3$), axial torsion ($q_4$), and gripper angle ($q_7$) in data-driven calibration.}
    \label{fig: data-driven motion profile}
\end{figure}

The instrument was randomly moved for 10 cycles in a cube of 40 mm in length that travels along the insertion axis of the PSM. A compressible foam was placed between the grippers and the gripper angle was randomly changed between 0 (firm grip) and 9 (loose grip) degrees. Above 9 degrees, the foam was loosely held in place and could fall so it was avoided. The motion profiles for the insertion axis ($q_3$), the axial torsion ($q_4$), and gripper angle ($q_7$) are shown in Fig. \ref{fig: data-driven motion profile}. The first 6 cycles were used for training, the cycles 7 and 8 were used for validation, and the last two cycles were used for testing. The ATI sensor and the optical force sensor were sampled at 1500 Hz to ensure low latency.
\begin{figure}[h]
    \centering
    \includegraphics[width=0.9\linewidth]{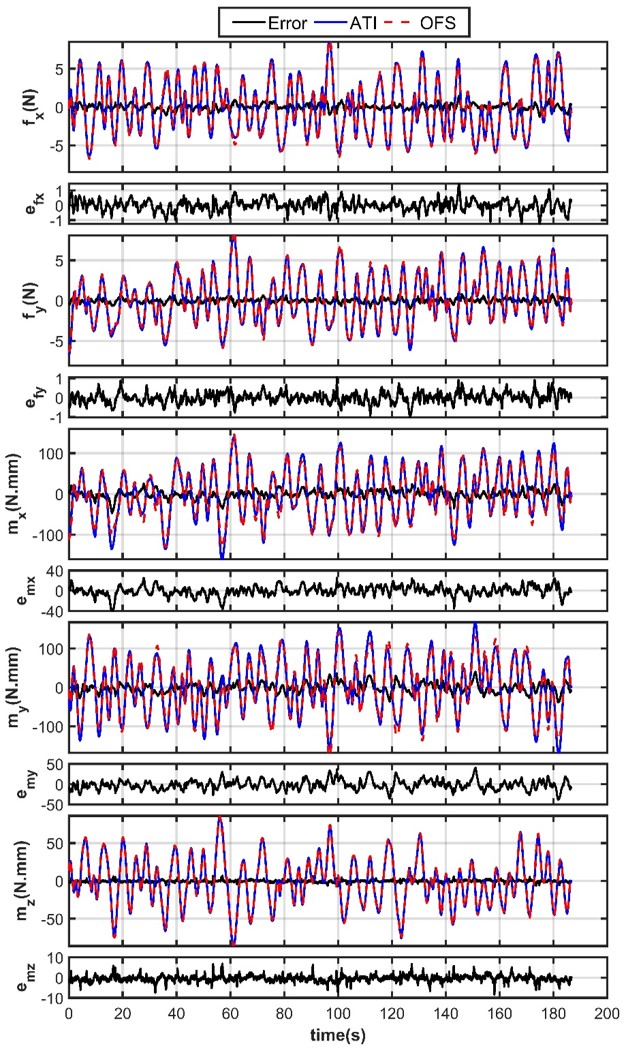}
    \caption{Force applied at the distal end of the instrument's shaft vs the calibrated optical force sensor mounted onto its proximal shaft - data-driven.}
    \label{fig: data-driven calibration}
\end{figure}

MATLAB's FitNet nonlinear regression tool was used for training different neural network architectures. The collected data was sub-sampled to 100 Hz to speed up the training. With the sub-sampled data, the training time reduces to less than a minute without a significant increase in the Mean-Square-Error (MSE). It was observed that a shallow network with fully connected layers cannot accurately resolve the axial force data. Because the axial force component had a big contribution to the MSE of the multi-axis regression, it was removed from the calibration set and the network was trained on the lateral forces and moments and the axial torque about the shaft axis. A 2-layer neural network, with 5 nodes in the hidden layer and 5 nodes in the output layer, was found to fit the validation set without overfitting the training data. The input layer is a 15$\times$1 vector as given in (\ref{eq: nn_input}) where $q_3$ and $q_4$ are the PSM's insertion in mm and axial torsion in rad., respectively, $m_g$ is the jaw effort in N$\cdot$mm, and $n_i$ are the normalized transducers signals of channels 1 to 6 of the optical force sensor.
\begin{equation}
    x_{in} = \begin{bmatrix} q_3,q_4,m_g,n_1, \cdots, n_6, n_1^2, \cdots, n_6^2 \end{bmatrix}^{T}
    \label{eq: nn_input}
\end{equation}

The jaw effort was added to compensate for the induced forces in the shaft when the instrument firmly grasps an object. It was observed that the forces due to the wrist maneuvers are lower than the calibration accuracy (see \ref{sec: wristManeuver}); therefore, the wrist angles were excluded from the input vector.

Figure \ref{fig: data-driven calibration} shows the data-driven calibration results in all the DoFs, except the axial force, for the validation (0 - 93 s) and test (93 - 186 s) sets. It shows that the calibrated sensor closely resolves the forces measured by the reference sensor.  It is important to note that, in the data-driven calibration, the effect of the jaw effort on the sensor readings is compensated by including the jaw effort in the network's input vector. It is much more complex to model and to compensate for in the model-based calibration. The rms and normalized-root-mean-square-deviation (NRMSD) of the error over the validation and test data can be calculated as an index of the sensor's repeatability and its calibration accuracy:
\begin{equation}
    NRMSD_i = \frac{\sqrt{\frac{1}{m}\sum\limits_{n=1}^m\left[\hat{f}_{ni}-f_{ni}\right]^2}}{f_{i,max}-f_{i,min}},
\end{equation}
where $m$ is the number of measurement points in each set, $i$ is the axis index, following the same sequence of forces and moments as in the wrench vector, $\hat{f}_{ni}$ are the calibration outputs and $f_{ni}$ are the reference (ATI) measurements. The $R^2$ values provide a measure of the sensor's linearity and hysteresis. The sensor's performance was quantified in different axes and is presented in Table \ref{T1}.
\begin{table}[h]
\renewcommand{\arraystretch}{1.3}
\centering
\caption{Data-driven Calibration characteristics of the optical force sensor: Range, repeatability (rms Error - $\sigma_i$), $NRMSD$, and $R^2$ \label{T1}}
\begin{tabular}{lcccccc}
\hline
Axis & $f_x$ & $f_y$ & $m_x$ & $m_y$ & $m_z$\\
Unit & N & N & N$\cdot$mm & N$\cdot$mm & N$\cdot$mm \\ 
$i$   & 1 & 2 & 4 & 5 & 6\\
\hline
Range & $\pm 9$ & $\pm 9$ & $\pm 160$ & $\pm 160$ & $\pm 100$ \\
$\sigma_i$ & 0.38 & 0.30 & 9.43 & 12.51 & 2.15 \\
$NRMSD_i$\small{(\%)}& 0.80 & 1.02 & 0.92 & 0.95 & 0.21 \\
$R^2_i$ (dmls) & 0.98 & 0.98 & 0.97 & 0.97 & 0.99 \\
\hline
\end{tabular}
\end{table}

\section{Design Evaluation}
\subsection{Overcoat Test}
\begin{figure}[h]
    \centering
    \includegraphics[width=0.9\linewidth]{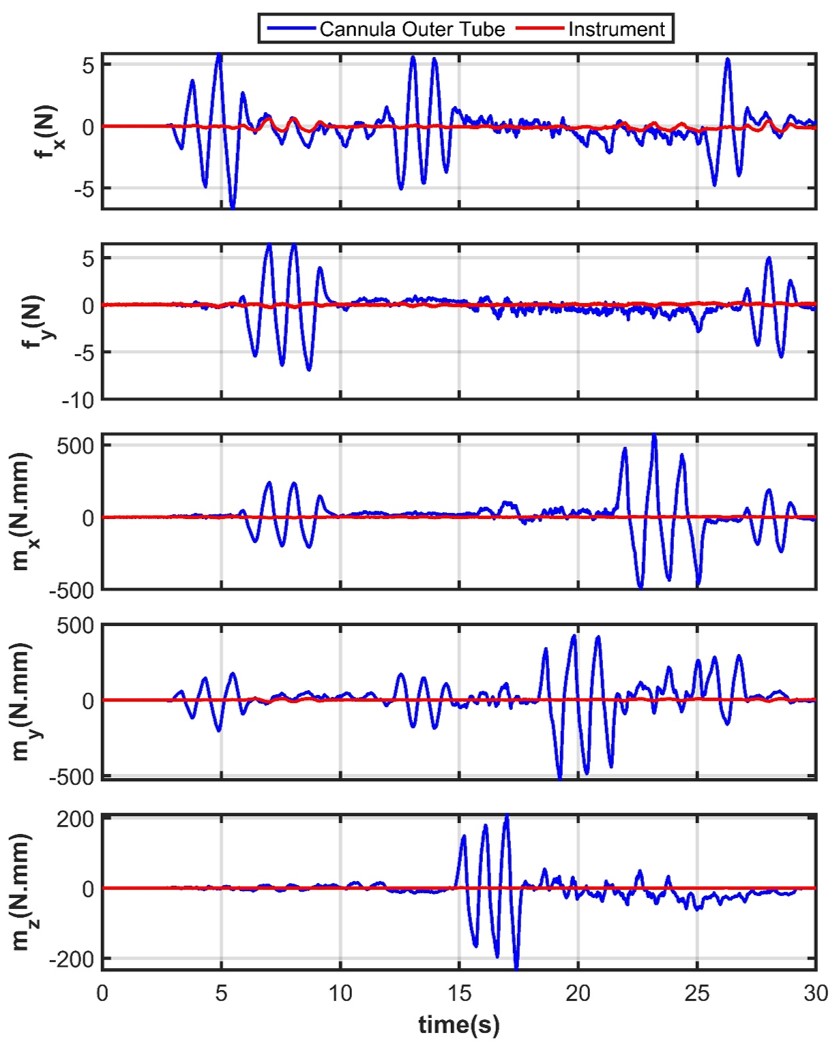}
    \caption{Overcoat test - The ATI sensor was attached to the outer tube of the cannula and the sensor signals are resolved using the calibration matrix to identify the effect of the body wall forces.}
    \label{fig: overcoat}
\end{figure}

The overcoat test was performed to evaluate the proposed cannula design in filtering the forces applied to its outer tube and isolating the load path from the instrument. The ATI Nano43 F/T sensor was attached to the cannula's outer tube, and sequences of forces and moments were applied to it. The normalized signals of the OFS' sensing modules and the ATI measurements were recorded. The NN calibration pipeline was used to resolve the wrench vector. As shown in Fig. \ref{fig: overcoat}, despite the relatively large forces and moments applied to the cannula's outer tube, the optical force sensor picks up minor oscillations in the resolved wrench vector. Hence, the cannula's two-layer design can mechanically filter out the body forces from the sensor readings. 
\begin{figure}[h]
    \centering
    \includegraphics[width=0.9\linewidth]{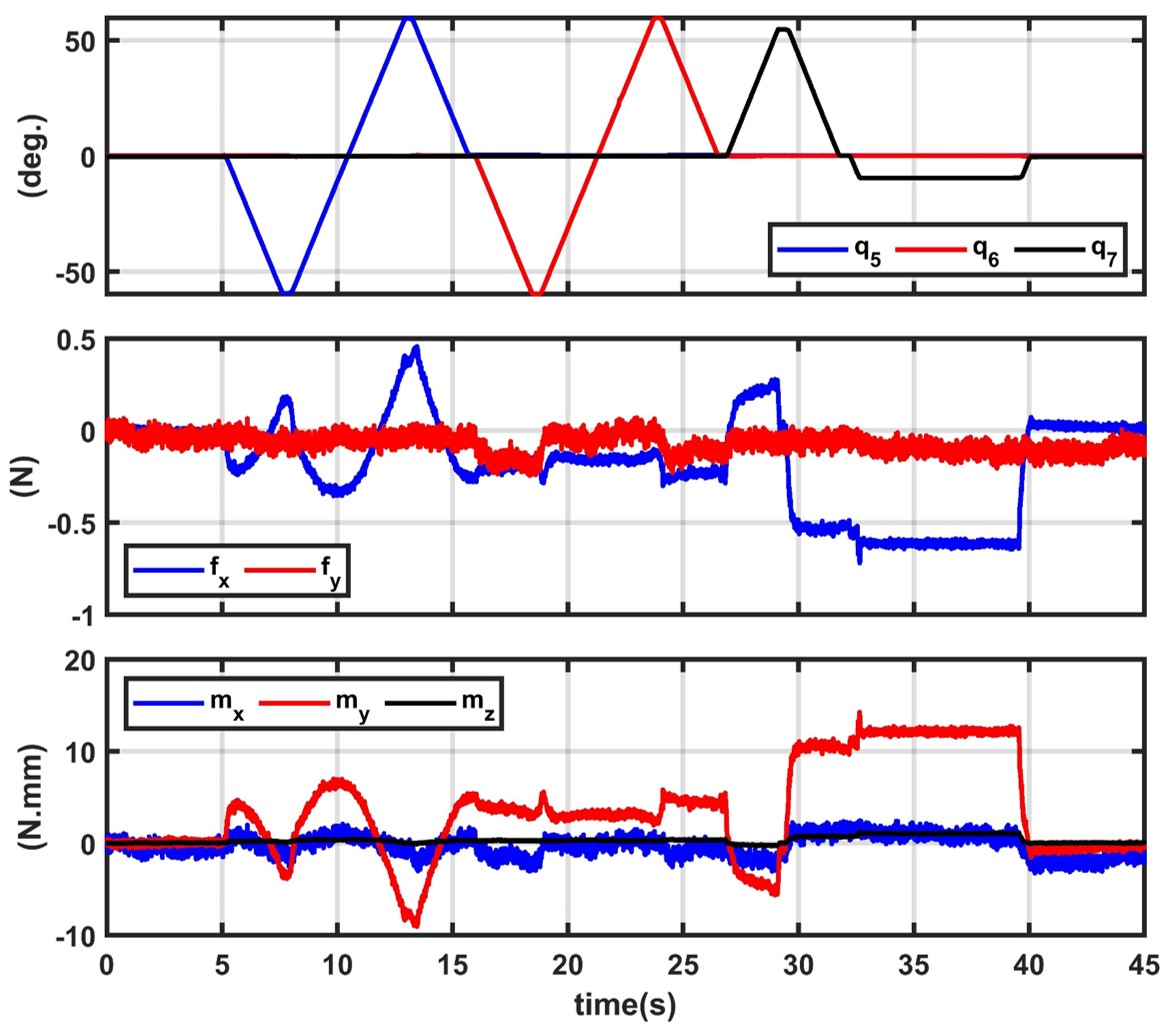}
    \caption{Wrist maneuver test - The instrument's wrist was moved within its mechanical range and the induced forces in the instrument shaft was estimated using the data-driven calibration pipeline.}
    \label{fig: wrist maneuver}
\end{figure}

\subsection{Wrist Maneuver Test}\label{sec: wristManeuver}
The wrist maneuver test was performed to evaluate the effect of wrist motions on forces generated in the instrument shaft. The wrist was sequentially moved between its mechanical limits in the pitch ($q_5$) and yaw ($q_6$) axes, and the gripper was fully opened and closed ($q_7$) as shown in the top plot of Fig. \ref{fig: wrist maneuver}. The gripper was commanded to -10$^\circ$ to generate a grasping force in the closed state. 
The two bottom plots show the forces and moments in the instrument shaft measured by the optical force sensor and the data-driven calibration pipeline (see section \ref{sec: data-driven calibration}). 
The plots show that the wrist motions and the gripper forces have minimal effect on the $f_y$, $m_x$, and $m_z$. However, they have a noticeable effect on the $f_x$, and $m_y$ components. Considering the range and the RMS values in Table \ref{T1}, the contribution of the wrist maneuvers and grasping forces on the measurements are within the 2$\sigma_i$ error margin. Without having the jaw effort ($m_g$) in the network inputs, the errors due to the wrist actuation could be as large as 20$\sigma_i$.

\section{Discussion and Conclusion}
In this paper, we proposed a novel 6-axis optical force sensor mounted onto the proximal shaft of a da Vinci EndoWrist instrument for measuring the wrench vector applied to its distal end. The optical force sensor has an active and a passive component. The active component has the power conditioning and digital electronics and six sensing modules. Each sensing module has a LED and a bicell placed inline, and a collocated signal conditioning board. The passive component has the slits in alternating orientations, aligned with the gap between the two active areas of the bicells, for light modulation. The careful electronics design provides a very high-resolution slit displacement measurement. 

The first prototype is compact and fits in a cube of 50~mm in length. It weighs approximately 150 grams, which is close to the weight of the ProGrasp instrument. A balancing slider with dummy weights is used in the parallelogram mechanism design of the da Vinci classic system for gravity compensation of the drive train and the instrument. The weights were adjusted to avoid the instrument dropping when the sensor is mounted and the drives are off. With the improvements in the next prototypes, the sensor's weight and size can be further reduced. 

The sensor has no structural flexure making it not breakable due to overload. The sensor’s components are not in the load path, and they do not require special provisions for mounting onto the structure. Therefore, it allows easy integration into a RAMIS systems. A higher number of sensing modules can be used for redundancy and error detection. Additionally, they can be placed in alternate configurations for improved accuracy and resolution. The signal conditioning electronics has a high-resolution ADC with a built-in programmable gain amplifier that can be used for dynamic adjustment of the transducer's range and resolution. The low latency and high data-rate reported in Section \ref{sec: sensor} make the sensor a good fit for control applications. With the onboard processor, only four wires (two for power and two for RS485 half-duplex communication link) are needed to interface with the sensor, which makes the wiring management simple and minimizes the cable loads in arm maneuvers. Lastly, the signal conditioning circuit has a high bandwidth of 500 Hz that allows using the sensor for vibration detection \cite{Hadi2014}, and vibrotactile haptic perception, as proposed by Kuchenbecker {\em et al.} \cite{Kuchenbecker2010}.

Despite the advantages above, the sensor has a few limitations: 1) The current design can be mounted onto a shaft of 8.3 mm in diameter. One aspect of design improvement is making the mounting interface adaptable to different shapes and diameters. 2) The sensor cannot be mounted onto the instrument before the instrument is attached to the classic, S, and SI da Vinci robots. It is because in the above series, the instrument clips onto a sterile adapter that intersects with the sensor's envelope. However, this is not a limitation in the X, and XI series because the mounting interface is different. 3) Mechanical references can be added for a more accurate positioning and repeatable sensor installation which could eliminate the need for re-calibration after every sensor installation onto the PSM. 4) The current sensor needs another reference sensor and the calibration setup in section \ref{sec: calibration} for calibration. In future work, we plan to explore other calibration methods that do not rely upon an external sensor, e.g., using the IMU and the inertial parameters of a known payload, or by payload estimation.

In addition to the force sensor, a modified cannula design was presented. The new design has two tubes. The outer tube isolates the load path from the instrument. The inner tube is attached through a 3-arm leaf spring to the base of the cannula to allow instrument bending due to forces at its distal end. The effective stiffness at the tip of the cannula's inner tube can be adjusted by rotating the inner tube and consequently changing the arm of the leaf springs. It is to avoid the closing of the 1.5 mm gap between the inner and the outer tube at the maximum lateral forces and moments. The proposed concept increased the OD of the cannula by 3 mm. This idea can be easily applied to the standard da Vinci cannula's currently used in clinical systems as well as the AirSeal\textregistered~access ports from CONMED \cite{AirSeal} for reduced friction.

A mathematical model was developed to capture the changes in the instrument's bending behavior as it penetrates into the cannula. It was used for a model-based calibration, and the results showed that the model captures the dynamics of the varying boundary condition. The model combined with the validated sensor model presented in \cite{Hadi2020-TRM} can be used for evaluating the design trade-offs and optimizations. The limitations of the model-based calibration were discussed. It was compared with a data-driven approach comprising a shallow neural network of one hidden layer and one output layer. In particular, the data-driven calibration covers the scenarios in which the developed model is not valid (see Fig. \ref{fig: model-based valid}), and it can compensate for the effect of the grasping force on the measurements that is complex to model.

A shallow NN architecture was used to avoid overfitting to the training data and minimize the computation cost. The data-driven calibration results showed an accuracy of close to 10\% in the lateral forces, which is the human's JND over the range of 0.5-200 N as explained in section \ref{sec: introduction}. As expected, the best accuracy ($\sim$6\%) was obtained in the axial torque. It is because the sensor configuration is most sensitive to the axial torque \cite{Hadi2020-TRM} and it is not affected by the changes in the instrument support (see (\ref{eq: bendModel}) and (\ref{eq: varyBoundCond})). Despite the high-resolution displacement measurement that the sensor provides, it failed to closely resolve the axial force. Design improvements such as using the AirSeal\textregistered~access port, and adding a Teflon coating or bronze bushings at the tip of the cannula's inner tube can improve the sensor performance in the axial direction by reducing the friction. Moreover, in the future, we will focus on cascading a supervised learning calibration designed to reconstruct the axial force components without degrading the sensor performance in the other DoFs.

The overcoat test and the wrist maneuver test were conducted to further evaluate the proposed sensing approach. The former showed that the modified cannula can properly filter the body wall forces from the measurements. The latter verified that the grasping forces have a more dominant effect on the measurement accuracy compared to the wrist actuation in the pitch and yaw axes.

%

\section*{Acknowledgment}
Amir Hossein Hadi Hosseinabadi gratefully acknowledges scholarship support from the NSERC Canada Graduate Scholarship-Doctoral program. Professor Salcudean gratefully acknowledges infrastructure support from CFI and funding support from NSERC and the Charles Laszlo Chair in Biomedical Engineering. The authors would also like to acknowledge help from the project electronics consultant Mr. Gerald F. Cummings and the co-op student Mr. David G. Black. The authors would also like to acknowledge Intuitive Surgical for initial discussions (Dr. Simon DiMaio) and providing the PSM used in this study.

\ifCLASSOPTIONcaptionsoff
  \newpage
\fi



%
\bibliographystyle{IEEEtran}
\bibliography{main}

%

\vspace{-1cm}
\begin{IEEEbiography}[{\includegraphics[width=1in,height=1.25in,clip,keepaspectratio]{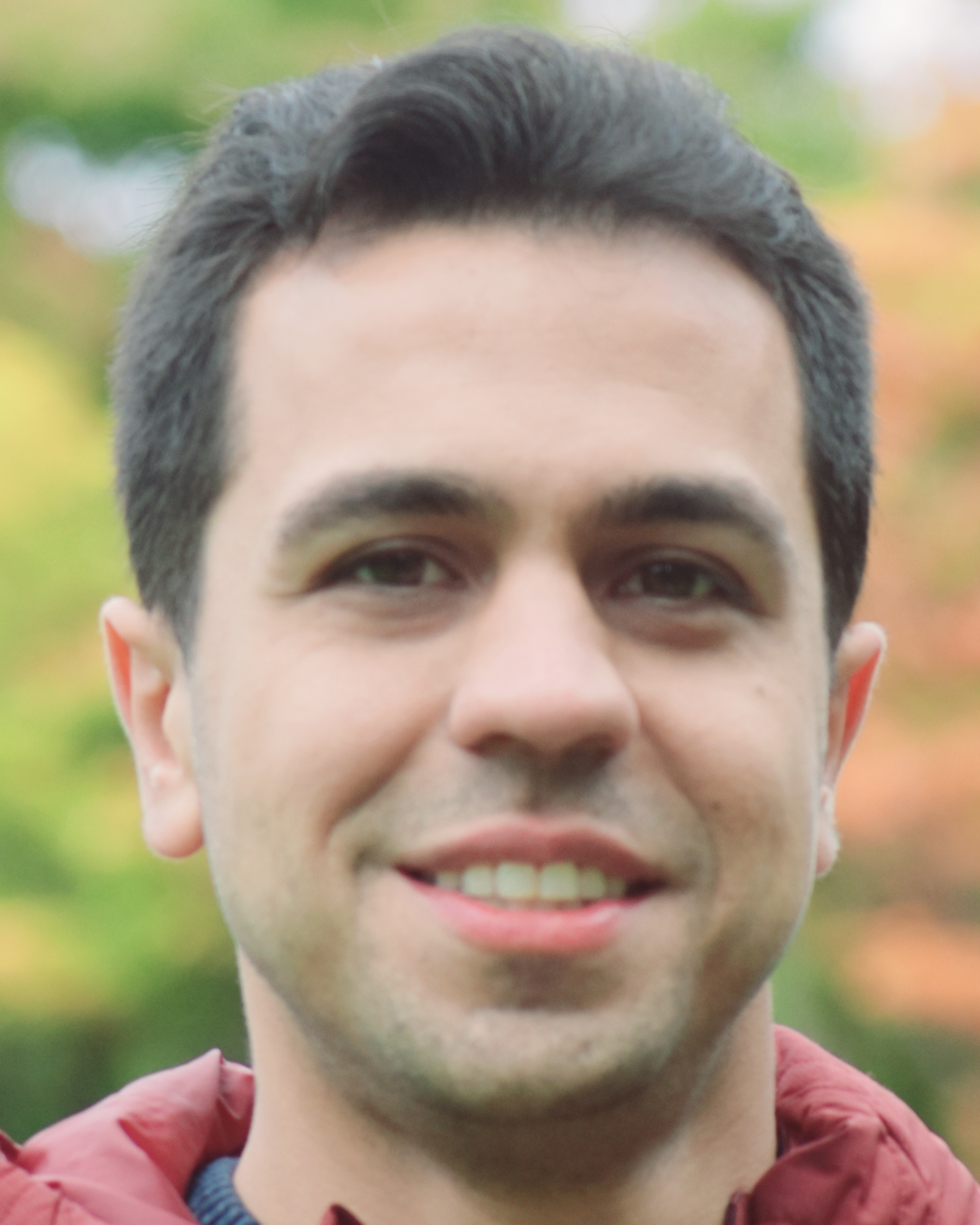}}]{Amir Hossein Hadi Hosseinabadi} was born in Esfahan, Iran in 1988. He received the BSc and MASc degrees in mechanical engineering in 2011 and 2013 from the Sharif University of Technology, Tehran, Iran, and the University of British Columbia (UBC), Vancouver, Canada, respectively. He is currently a PhD candidate in electrical and computer engineering at UBC and a research assistant at the Robotics and Control Laboratory (RCL). From 2013-2020, he was a Robotics \& Control Engineer at Dynamic Attractions, Port Coquitlam, Canada. He completed an internship at Microsoft, Redmond, WA, USA from December 2020 to February 2021 and is currently a Ph.D. research intern with Intuitive Surgical, Sunnyvale, CA, USA.
\end{IEEEbiography}
\vspace{-1cm}
\begin{IEEEbiography}[{\includegraphics[width=1in,height=1.25in,clip,keepaspectratio]{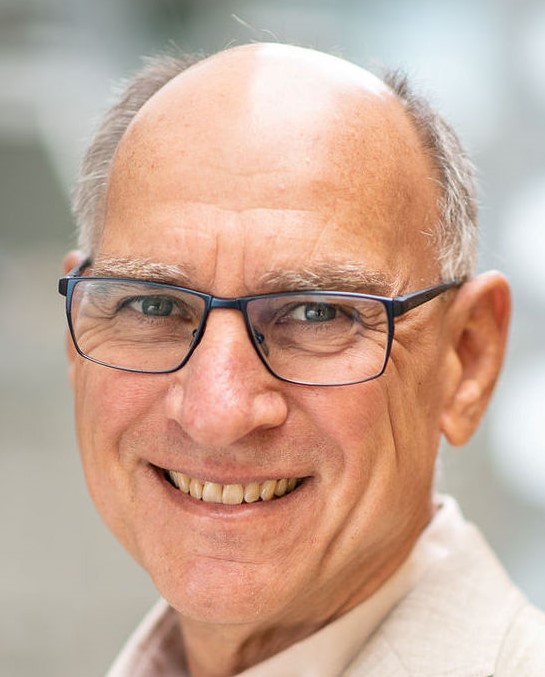}}]{Septimu E. Salcudean}was born in Cluj, Romania. He received the BEng (Hons.) and MEng degrees in from McGill University, Montreal, Quebc, Canada in 1979 and 1981, respectively, and his PhD degree from the University of California, Berkeley, USA in 1986, all in electrical engineering.\\
He was a Research Staff Member at the IBM T.J. Watson Research Center from 1986 to 1989. He then joined the University of British Columbia (UBC) and currently is a Professor in the Department of Electrical and Computer Engineering, where he holds the C.A. Laszlo Chair in Biomedical Engineering and a Canada Research Chair. He has courtesy appointments with the UBC School of Biomedical Engineering and the Vancouver Prostate Centre. He has been a co-organizer of the Haptics Symposium, a Technical Editor and Senior Editor of the IEEE Transactions on Robotics and Automation, and on the program committees of the ICRA, MICCAI and IPCAI Conferences. He is currently on the steering committee of the IPCAI conference and on the Editorial Board of the International Journal of Robotics Research. He is a Fellow of the IEEE, a Fellow of MICCAI and of the Canadian Academy of Engineering.
\end{IEEEbiography}
\vfill\eject




\end{document}